\documentclass[11pt,a4paper]{article}

\usepackage[hyperref]{naaclhlt2019}
\usepackage[T1]{fontenc}
\usepackage{times}
\usepackage{latexsym}
\usepackage{url}
\usepackage[capitalise,nameinlink]{cleveref}
\usepackage{booktabs}
\usepackage{multirow}
\usepackage{graphicx}

\makeatletter
\g@addto@macro{\UrlBreaks}{\UrlOrds}
\makeatother

\newcommand*{\textcite}[1]{\citeauthor{#1}~\shortcite{#1}}
\newcommand*{\parencite}[1]{\cite{#1}}

\newcommand*\ClearAux[0]{$\mathrm{CLEAR}^\mathrm{aux}$}
\newcommand*\ClearMlpAux[0]{$\mathrm{CLEAR}_\mathrm{MLP}^\mathrm{aux}$}
\newcommand*\ClearCnnMlp[0]{$\mathrm{CLEAR}_\mathrm{CNN+MLP}$}
\newcommand*\ClearCnnMlpAux[0]{$\mathrm{CLEAR}_\mathrm{CNN+MLP}^\mathrm{aux}$}
\newcommand*\ClearNileTMRG[0]{$\mathrm{CLEAR}_\mathrm{NileTMRG}$}
\newcommand*\ClearSubtaskB[0]{$\mathrm{CLEAR}_\mathrm{Subtask\mbox-B}$}

\aclfinalcopy

\title{CLEARumor at SemEval-2019 Task 7:\\ConvoLving ELMo Against Rumors}
\author{
  Ipek Baris\textsuperscript{1,}\footnotemark[1]
  \and Lukas Schmelzeisen\textsuperscript{1,}\thanks{\ \ The first two authors contributed equally.}
  \and Steffen Staab\textsuperscript{1,\,2}\\
  \textsuperscript{1}Institute for Web Science and Technologies (WeST), University of Koblenz-Landau, Germany\\
  \textsuperscript{2}Web and Internet Science Group (WAIS), University of Southampton, United Kingdom\\
  \texttt{ibaris@uni-koblenz.de}, \texttt{lukas@uni-koblenz.de}, \texttt{staab@uni-koblenz.de}
}

\date{}

\begin{document}
\frenchspacing
\maketitle

\begin{abstract}
This paper describes our submission to SemEval-2019 Task 7: RumourEval: Determining Rumor Veracity and Support for Rumors.
We participated in both subtasks.
The goal of subtask A is to classify the type of interaction between a rumorous social media post and a reply post as support, query, deny, or comment.
The goal of subtask B is to predict the veracity of a given rumor. 
For subtask A, we implement a CNN-based neural architecture using ELMo embeddings of post text combined with auxiliary features and achieve a F\textsubscript{1}-score of $44.6\%$.
For subtask B, we employ a MLP neural network leveraging our estimates for subtask A and achieve a F\textsubscript{1}-score of $30.1\%$ (second place in the competition).
We provide results and analysis of our system performance and present ablation experiments.    
\end{abstract}

\section{Introduction}
\label{sec:introduction}

Online social media has changed the way of communicating and disseminating media content and opinions, but also paved the way for spreading false or unverified rumors.  

\begin{figure}[t]
  \centering
  \includegraphics[width=0.75\columnwidth]{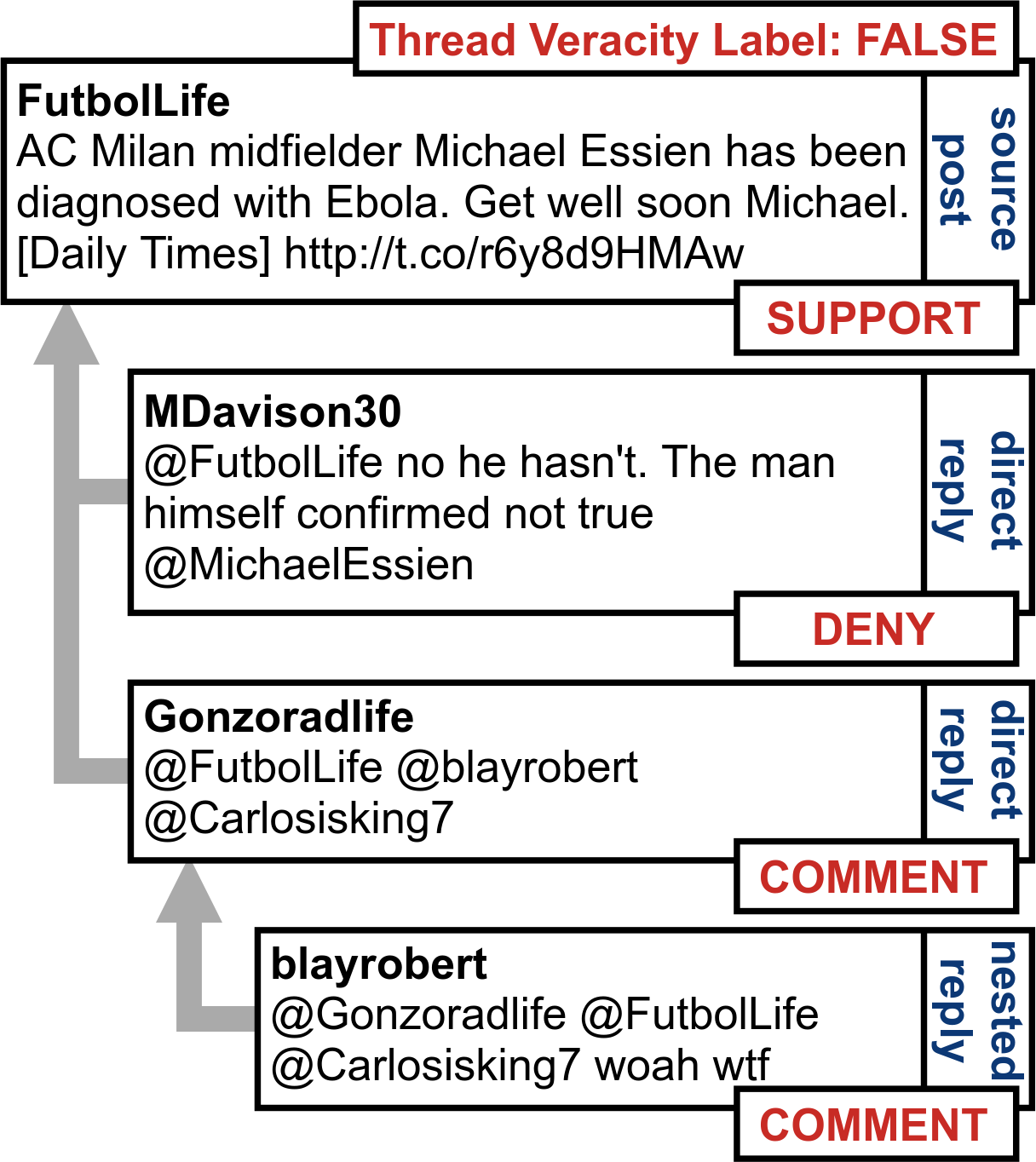}
  \caption{An example Twitter thread from the training dataset with SDQC labels for each post and a veracity label for the thread's source post.
    Any post that does not reply to another is a \emph{source post}.
    Reply posts can be \emph{direct replies} (replies to a source post) or \emph{nested replies} (replies that reply to another reply post).
    A \emph{thread} is the set containing a source post and all its reply posts.}
  \label{fig:example}
\end{figure}

RumourEval 2019 \parencite{conf/semeval/GorrellBDKLZ19} provides a dataset of labelled threads from Twitter and Reddit where each source post mentions a rumor.
Subtask A (SDQC) consists of deciding for each post in a thread whether it is in a \emph{support}, \emph{deny}, \emph{query}, or \emph{comment} relation to the rumor.
The goal of subtask B (Verification) it to classify the veracity of the rumor as \emph{true}, \emph{false}, or \emph{unverified}.
\Cref{fig:example} clarifies our terminology and the tasks.

Automated rumor classification is a challenging task as there is no definite evidence (e.g., authorized confirmation).
In its absence, stance analysis is a useful approach.
Systems that employ neural network architectures showed promising results in RumourEval 2017 \parencite{DBLP:conf/semeval/DerczynskiBLPHZ17}, with the LSTM-based sequential model of \textcite{DBLP:conf/semeval/X17a} performing best.

In this paper, we describe our approach CLEARumor (ConvoLving ELMo Against Rumors) for solving both subtasks and provide empirical results and ablation experiments of our architecture.
We make our PyTorch-based implementation and trained models publicly available\footnote{\url{https://github.com/Institute-Web-Science-and-Technologies/CLEARumor}}.

\section{System Description}
\label{sec:system-description}

After preprocessing the post text (\cref{subsec:preprocessing}) and embedding it with ELMo (\cref{subsec:elmo}), our architecture for subtask A (\cref{subsec:sdqc-architecture}) passes the embedded text through a convolutional neural network (CNN) block, adds auxiliary features, and uses  a multilayer perceptron (MLP) block for estimating class membership.
These estimates are combined with further auxiliary features and fed into an MLP block for the classification for subtask B (\cref{subsec:verif-architecture}).

\subsection{Preprocessing}
\label{subsec:preprocessing}

For preprocessing, we rely mostly on Erika Varis Doggett's tokenizer for Twitter and Reddit\footnote{\url{https://github.com/erikavaris/tokenizer}}, with which we strip away all user handles (e.g., ``@FutbolLife''), remove the number sign in front of hash tags (e.g., ``\#Ebola'' becomes ``Ebola''), remove URLs, and limit repetitions of the same character to at most three times (e.g., ``heeeeey'' becomes ``heeey'').
We further decided to lowercase all text, which resulted in improved performance over mixed case in initial experiments.
Last, all posts are truncated after 32 tokens\footnote{Only 10 out of the total 6634 Twitter posts are longer than this, while a few Reddit posts are up to 1,000 tokens long which would result in very impractical batch sizes.}.

\subsection{ELMo Embeddings}
\label{subsec:elmo}

The task of word embedding is to represent each word in a given sentence by a vector, which among other things allows for encoding words at the input layer in a neural network architecture.
Traditional embedding methods such as word2vec \parencite{DBLP:conf/nips/MikolovSCCD13} or GloVe \parencite{DBLP:conf/emnlp/PenningtonSM14} work independently of context and always map the same word to the same vector.

In contrast, ELMo \parencite{DBLP:conf/naacl/PetersNIGCLZ18} is a recent embedding approach based on bidirectional LSTM networks that considers the context a word occurs in and is thereby able to address certain linguistic peculiarities, e.g., that the same word can have different meanings depending on its context.
Further, ELMo incorporates subword units and is thereby able to represent words not seen during training successfully, an important benefit for the social media domain, where users frequently misspell existing words or introduce new ones.
Formally, given a sequence of words $w_1 w_2 \dots w_n$ ELMo represents the $k$-th word $w_k$ as
\begin{equation}
  \mathbf{ELMo}^\mathrm{task}_k = \gamma^\mathrm{task} \sum_{j=0}^L s^\mathrm{task}_j \mathbf{h}_{k, j},
\end{equation}
where $L$ gives the number of internal layers that were used to train ELMo, $\mathbf{h}_{k, j}$ is the contextual vector representation of layer $j$ for word $k$, and $\gamma^\mathrm{task}$ and the $s^\mathrm{task}_j$ are scalars that can be tuned specifically for the task at hand.

We report results for the pretrained model \texttt{elmo\_2x4096\_512\_2048cnn\_2xhighway\allowbreak\_5.5B}\footnote{\url{https://allennlp.org/elmo}} for which $L=2$ and which outputs $1024$-dimensional embedding vectors (but didn't notice drastic improvements over the much smaller models).
ELMo allows us to fine-tune $\gamma^\mathrm{task}$, $s^\mathrm{task}_j$, and even the $\mathbf{h}_{k,j}$ by backpropagating gradients to them, but we decided against this, because the RumourEval dataset is very small (cf. \cref{tab:dataset}) adjusting these weights can quickly lead to overfitting, and keeping the weights constants allows us to precompute and store all ELMo embeddings once before the training process which results in a major boost in performance.

\subsection{Subtask A}
\label{subsec:sdqc-architecture}

\begin{figure}
  \centering
  \includegraphics[width=0.75\columnwidth]{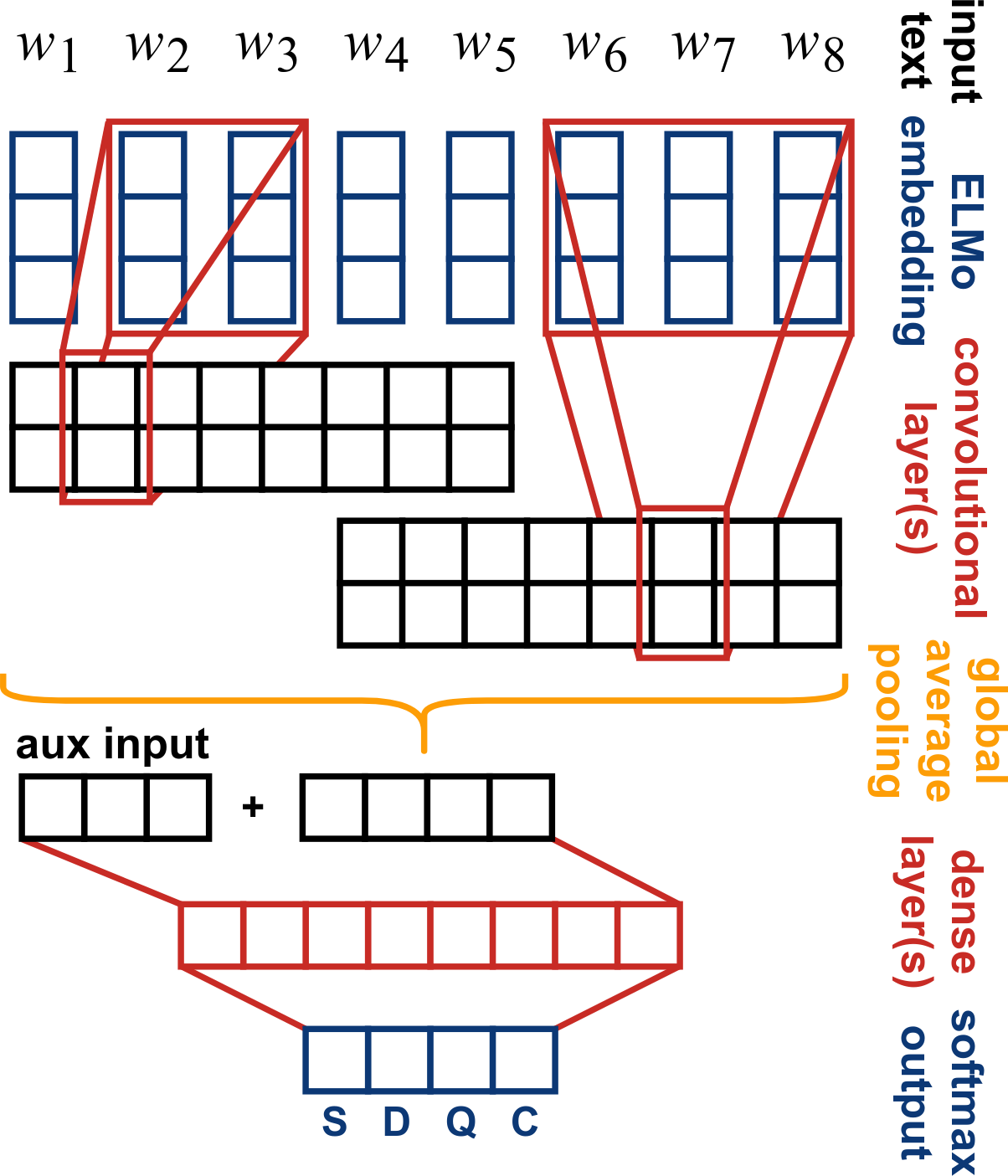}
  \caption{CLEARumor architecture for subtask A.}
  \label{fig:subtaska-architecture}
\end{figure}

Our architecture for subtask A is visualized in \cref{fig:subtaska-architecture}.
First, the tokenized text of the post that is to be classified as either support, deny, query, or comment is represented with an ELMo embedding.
Next, the embedded text is fed into $L_\mathrm{conv}$-many convolutional layers.
Here, a single convolutional layers consists of multiple 1D-convolution operations with a set of different kernel sizes $S$, each mapping onto $C$ convolutional channels, which are then concatenated along the channel axis.
Each convolution operation is batch normalized \parencite{DBLP:conf/icml/IoffeS15} after a ReLU activation.
To maintain an equal sequence length, sequences are padded with zero vectors.
The resulting sequence representation is transformed into a single $\vert S \vert \cdot C$-dimensional vector via global average pooling.
This sequence vector is concatenated with a vector of auxiliary features that encodes meta information about the post under classification (detailed in the next paragraph).
Following is a stack of $L_\mathrm{dense}$-many dense layers, for which dropout-regularization \parencite{DBLP:journals/jmlr/SrivastavaHKSS14} is performed after ReLU activation.
Finally, a single linear layer that is softmax-activated yields the four estimates of class membership.
Parameters are optimized using Adam \parencite{DBLP:journals/corr/KingmaB14} and a cross-entropy loss.

We use the following auxiliary features:
(1)~a two-dimensional Boolean vector encoding whether the post is from Twitter or Reddit;
(2)~a five-dimensional real-valued vector encoding meta-information for the post author: whether the user is verified or not, the number of followers they have, the number of accounts they follow themselves, and a ratio of the latter two numbers\footnote{We use min-max scaling based on the training data for these features. For Reddit the respective concepts don't exist and a vector of zeros is used instead.};
(3)~the cosine similarity of the averaged ELMo embeddings of the post under classification to those of the thread's source post (defined to be $1$ for source posts); and
(4)~a three-dimensional Boolean vector encoding whether the post is a source post, a direct reply, or a nested reply.

As hyperparameters, we employ a learning rate of $10^{-3}$, a batch size of $512$, and train for $100$ epochs.
In our loss function, we weigh the estimates of support, deny, and query equally but that of comment at only a fifth of the strength because of the imbalance of the dataset.
We add L2-regularization with a weight of $10^{-2}$.
In our reported results we use $L_\mathrm{conv} = 1$ convolutional layer, with kernel sizes $S = \{2,3\}$ each mapping into $C=64$ channels, after which follow $L_\mathrm{dense} = 3$ dense layers with $128$ hidden units each and a dropout of $0.5$.

\subsection{Subtask B}
\label{subsec:verif-architecture}

For subtask B we build a single feature vector that we feed into a MLP classifier. We reuse all the auxiliary features from subtask A except the last two, because all posts under classification in subtask B are source posts.
We further add the following features:
(1) a two-dimensional Boolean vector encoding whether media (an image or a URL) is attached to the post,
(2) the upvote-to-downvote ratio of the post for Reddit (manually set to $0.5$ for Twitter),
(3) a two-dimensional real-valued vector encoding which fraction of the thread's posts are direct replies and which fraction are nested replies,
(4) the averaged support, deny, and query probability estimates from subtask A averaged over all posts in the thread.
Similarly to subtask A, this feature vector is fed into a stack of $L_\mathrm{dense}$-many dense layers with dropout-regularization \parencite{DBLP:journals/jmlr/SrivastavaHKSS14} after ReLU activation, after which a single softmax-activated linear layer yields estimates for the three classes true, false, or unverified.

Our model was trained with a learning rate of $10^{-3}$ and a batch size of $128$ for $5000$ epochs.
In our loss calculation, we weigh the unverified class at $0.3$ of the strength of the other two, and add a L2-regularization weight of $10^{-2}$.
We used $L_\mathrm{dense} = 2$ dense layers with $512$ hidden units each and a dropout of $0.25$.

\section{Evaluation}
\label{sec:evaluation}

\begin{table}[b!]
  \centering
  \addtolength{\tabcolsep}{-2pt}    
  \begin{tabular}{@{}ll rrrr r@{}}
    \toprule
    \multicolumn{2}{@{}l}{\bf Subtask A} & \multicolumn{1}{c}{\bf S} & \multicolumn{1}{c}{\bf D} & \multicolumn{1}{c}{\bf Q} & \multicolumn{1}{c}{\bf C} & \multicolumn{1}{c}{$\bf\Sigma$} \\
    \midrule
    \multirow{2}{*}{\bf Train}  & Twitter &  910 & 344 & 358 & 2907 & \multirow{2}{*}{\bf 5217} \\
                                & Reddit  &  15  & 34  & 37  & 612  & \\
    \midrule
    \multirow{2}{*}{\bf Dev}    & Twitter &  94  & 71  & 106 & 778  & \multirow{2}{*}{\bf 1485} \\
                                & Reddit  &  8   & 11  & 14  & 403  & \\
    \midrule
    \multirow{2}{*}{\bf Test}   & Twitter &  141 & 92  & 62  & 771  & \multirow{2}{*}{\bf 1827} \\
                                & Reddit  &  16  & 9   & 31  & 705  & \\
    \midrule
    {$\bf\Sigma$}               &         &  {\bf 1184} & {\bf 561} & {\bf 608} & {\bf 6176} & {\bf 8529} \\
    \bottomrule
  \end{tabular}
  \addtolength{\tabcolsep}{2pt}    
  
  \vspace{0.25cm}
  
  \addtolength{\tabcolsep}{2pt}    
  \begin{tabular}{@{}ll rrr r@{}}
    \toprule
    \multicolumn{2}{@{}l}{\bf Subtask B} & \multicolumn{1}{c}{\bf T} & \multicolumn{1}{c}{\bf F} & \multicolumn{1}{c}{\bf U} & \multicolumn{1}{c}{$\bf\Sigma$} \\
    \midrule
    \multirow{2}{*}{\bf Train}  & Twitter &  137 & 62  & 98  & \multirow{2}{*}{\bf 327} \\
                                & Reddit  &  7   & 17  & 6  & \\
    \midrule
    \multirow{2}{*}{\bf Dev}    & Twitter &  8   & 12  & 8   & \multirow{2}{*}{\bf 38} \\
                                & Reddit  &  2   & 7   & 1   & \\
    \midrule
    \multirow{2}{*}{\bf Test}   & Twitter &  22  & 30  & 4   & \multirow{2}{*}{\bf 81} \\
                                & Reddit  &  9   & 10  & 6   & \\
    \midrule
    {$\bf\Sigma$}               &         & {\bf 185} & {\bf 138} & {\bf 133} & {\bf 456} \\
    \bottomrule
  \end{tabular}
  \addtolength{\tabcolsep}{-2pt}    
  \caption{Number of labelled instances for both subtasks of the RumourEval 2019 dataset broken down into (1)~class frequencies, per (2)~social media platform, and (3)~training, development, and test dataset.}
  \label{tab:dataset}
\end{table}

\begin{table*}[t]
  \centering
  \addtolength{\tabcolsep}{-2.25pt}
  \begin{tabular}{@{}l r@{}l r@{}l r@{}lr@{}lr@{}lr@{}l r@{}l@{}}
    \toprule
    \multirow{2}{*}{\bf Subtask A}& \multicolumn{2}{c}{\bf Dev} & \multicolumn{10}{c}{\bf Test} & \multicolumn{2}{c}{\bf CV} \\
    \cmidrule(lr){2-3} \cmidrule(lr){4-13} \cmidrule(lr){14-15}
    & \multicolumn{2}{c}{\bf Macro-F\textsubscript{1}} & \multicolumn{2}{c}{\bf Macro-F\textsubscript{1}} & \multicolumn{2}{c}{\bf S-F\textsubscript{1}} & \multicolumn{2}{c}{\bf D-F\textsubscript{1}} & \multicolumn{2}{c}{\bf Q-F\textsubscript{1}} & \multicolumn{2}{c}{\bf C-F\textsubscript{1}} & \multicolumn{2}{c}{\bf Macro-F\textsubscript{1}} \\
    \midrule
    Always Comment  &     $22.1$ &         &     $22.3$ &         &      $0.0$ &         &      $0.0$ &         &      $0.0$ &         &     $89.4$ &         & \multicolumn{2}{c}{---} \\
    Submitted       &     $41.3$ &         &     $37.4$ &         & {$\bf46.7$}&         &      $0.0$ &         &     $11.7$ &         & {$\bf91.2$}&         & \multicolumn{2}{c}{---}  \\
    \ClearAux       & {$\bf44.8$}&$\pm0.6$ &     $42.7$ &$\pm0.6$ &     $29.6$ &$\pm0.6$ & {$\bf17.8$}&$\pm2.4$ &     $43.9$ &$\pm1.0$ &     $79.5$ &$\pm1.3$ &     $47.1$ &$\pm4.5$ \\
    \ClearMlpAux    &     $42.2$ &$\pm1.2$ &     $40.7$ &$\pm1.6$ &     $30.7$ &$\pm2.7$ &      $0.0$ &$\pm0.0$ & {$\bf51.6$}&$\pm3.2$ &     $80.5$ &$\pm2.7$ &     $44.7$ &$\pm4.2$ \\
    \ClearCnnMlp    &     $39.7$ &$\pm2.0$ &     $39.0$ &$\pm2.2$ &     $16.2$ &$\pm2.3$ &     $14.8$ &$\pm3.4$ &     $41.0$ &$\pm6.7$ &     $84.0$ &$\pm2.6$ &     $43.3$ &$\pm4.5$ \\
    \ClearCnnMlpAux &     $42.9$ &$\pm2.2$ & {$\bf44.6$}&$\pm2.6$ &     $34.6$ &$\pm3.7$ &     $15.4$ &$\pm3.1$ &     $42.2$ &$\pm8.3$ &     $86.1$ &$\pm1.1$ & {$\bf47.2$}&$\pm3.8$ \\ 
    \bottomrule
  \end{tabular}
  \addtolength{\tabcolsep}{2.25pt}
  \caption{Evaluation results for subtask A.
    All reported scores are multiplied by 100.
    We provide the macro-averaged F\textsubscript{1}-score for the development (Dev), the test (Test) datasets and for 10-fold cross validation (CV).
    For the test dataset, we further provide the individual F\textsubscript{1}-scores per class.
    ``Always Comment'' is a baseline predicting always the most common class.
    ``Submitted'' are the results we officially submitted to RumourEval 2019.
    For our CLEARumor architecture we provide multiple ablation experiments.
    \ClearCnnMlpAux\ is our full system, \ClearCnnMlp\ the same but without the auxiliary features, \ClearMlpAux\ instead uses no convolutional layers, and \ClearAux\ just concatenates averages ELMo embeddings with auxiliary features uses a single linear layer.}
  \label{tab:subtaska-results}
\end{table*}

\begin{table*}[t]
  \centering
  \addtolength{\tabcolsep}{-2pt}    
  \begin{tabular}{@{}l r@{}lr@{}l r@{}lr@{}l r@{}lr@{}l@{}}
    \toprule
    \multirow{2}{*}{\bf Subtask B}& \multicolumn{4}{c}{\bf Dev} & \multicolumn{4}{c}{\bf Test} & \multicolumn{4}{c}{\bf CV} \\
    \cmidrule(lr){2-5} \cmidrule(lr){6-9} \cmidrule(lr){10-13}
    & \multicolumn{2}{c}{\bf Macro-F\textsubscript{1}} & \multicolumn{2}{c}{\bf RMSE} & \multicolumn{2}{c}{\bf Macro-F\textsubscript{1}} & \multicolumn{2}{c}{\bf RMSE} & \multicolumn{2}{c}{\bf Macro-F\textsubscript{1}} & \multicolumn{2}{c}{\bf RMSE} \\
    \midrule
    Submitted       &     $41.7$ &         &     $0.743$ &           &     $28.6$ &         &     $0.764$ & & \multicolumn{2}{c}{---} & \multicolumn{2}{c}{---} \\
    \ClearSubtaskB  &     $35.4$ &$\pm0.5$ & {$\bf0.676$}&$\pm0.005$ & {$\bf30.1$}&$\pm0.8$ & {$\bf0.754$}&$\pm0.005$ & {$\bf26.7$}&$\pm13.4$ & {$\bf0.733$}&$\pm0.113$ \\
    \ClearNileTMRG  & {$\bf53.5$}&         &     $0.761$ &           &     $18.6$ &         &     $0.846$ & &  \multicolumn{2}{c}{---} & \multicolumn{2}{c}{---} \\
    \bottomrule
  \end{tabular}
  \addtolength{\tabcolsep}{2pt}
  \caption{Evaluation results for subtask B.
    We report F\textsubscript{1} (multiplied by 100) and RMSE (root mean squared error) scores for the development (Dev), the test (Test) datasets and for 10-fold cross validation (CV).
    \ClearSubtaskB\ is our subtask B architecture using the subtask A estimates from \ClearCnnMlpAux.
    \ClearNileTMRG\ uses the same estimates but computes task B results using the NileTMRG system \parencite{DBLP:conf/semeval/EnayetE17}.}
  \label{tab:subtaskb-results}
\end{table*}

The dataset of RumourEval 2019 is summarized in \cref{tab:dataset}.
Our results for subtask A and B are detailed in \cref{tab:subtaska-results} and \cref{tab:subtaskb-results}, respectively.

The reported results differ from our official submission, because we continued to tune hyperparameters afterwards.
We report results as trained on the training dataset and then evaluated on the development and test datasets, as provided by the RumourEval organizers.
Because neural network experiments are naturally nondeterministic \parencite{DBLP:journals/corr/abs-1803-09578} and we did indeed notice huge variances when retraining models, we report the mean and standard deviation over $10$ runs for each experiment.
Additionally, we report scores from a $10$-fold cross validation over the whole dataset.
Simple cross-validation would be inappropriate in our setting, because for example a split could result in the case where the same rumors occur in both the training and the test dataset which would allow a model to just memorize which posts are rumorous.
We ensure that this does not happen in our case, by keeping all posts belonging to the same rumor\footnote{For Twitter posts, the dataset contains rumor-topic labels for each thread, so we ensure that each topic only occurs in one fold. For Reddit posts, no labelling is available, so we can only ensure that all posts of a thread occur in the same fold.} in the same cross validation fold.
Note that scores on the organizer split and the cross validation are not directly comparable as different fractions of the whole dataset are used for training (\textasciitilde60-70\% for the organizer split and \textasciitilde90\% for 10-fold cross validation).

\section{Conclusion}
\label{sec:conclusion}

We have presented CLEARumor, our architecture for the RumourEval 2019 shared tasks.
In future we aim to generalize our approach, e.g., we currently use domain-specific features for characterizing the post author popularity, such as number of followers for Twitter, which are not available for all social media platforms.
Besides investigating how well our approach translates to other languages, we are interested in studying the results for other pretrained word representation approaches, e.g., BERT \parencite{DBLP:journals/corr/abs-1810-04805}.

\section*{Acknowledgments}

Ipek Baris works for the EU-Project Co-Inform which has received funding from the European Union's Horizon 2020 research and innovation programme under grant agreement No 770302.
Lukas Schmelzeisen is supported by the German Research Foundation (DFG) as part of the Open Argument Mining project (grant STA 572/18-1).

\bibliography{clearumor}
\bibliographystyle{acl_natbib}

\end{document}